\pdfoutput=1

\documentclass[11pt]{article}

\usepackage[]{ACL2023}

\usepackage{times}
\usepackage{latexsym}
\usepackage{bm}
\usepackage{amsmath}
\usepackage{amssymb}
\usepackage{graphicx}
\usepackage{booktabs}
\usepackage{multirow}
\usepackage{makecell}
\usepackage{bbding}
\usepackage{pifont}
\usepackage{enumitem}
\usepackage{hyperref}
\usepackage{subfigure}
\usepackage[T1]{fontenc}

\usepackage[utf8]{inputenc}

\usepackage{microtype}

\usepackage{inconsolata}

\title{Towards Better Graph-based Cross-document Relation Extraction via Non-bridge Entity Enhancement and Prediction Debiasing}

\author{Hao Yue\textsuperscript{1,2}, Shaopeng Lai\textsuperscript{1,2}, Chengyi Yang\textsuperscript{1,2}, Liang Zhang\textsuperscript{1,2}, Junfeng Yao\textsuperscript{1,2}, Jinsong Su\textsuperscript{1,2\thanks{*Corresponding author.}}\\
  \textsuperscript{1}School of Informatics,  Xiamen University \\
  \textsuperscript{2}Xiamen Key Laboratory of Intelligent Storage and Computing, School of Informatics, Xiamen University \\
  \texttt{\{yuehao,lzhang\}@stu.xmu.edu.cn},
  \texttt{laishaopeng.lsp@alibaba-inc.com},\\\texttt{yangchengyi031@gmail.com},
  \texttt{\{yao0010,jssu\}@xmu.edu.cn}
  }
\begin{document}
\maketitle
\begin{abstract}
Cross-document Relation Extraction aims to predict the relation between target entities located in different documents. 
In this regard, the dominant models commonly retain useful information for relation prediction via bridge entities, which allows the model to elaborately capture the intrinsic interdependence between target entities. However, these studies ignore the non-bridge entities, each of which co-occurs with only one target entity and offers the semantic association between target entities for relation prediction. Besides, the commonly-used dataset--CodRED contains substantial NA instances, leading to the prediction bias during inference. To address these issues, in this paper, we propose a novel graph-based cross-document RE model with non-bridge entity enhancement and prediction debiasing. Specifically, we use a unified entity graph to integrate numerous non-bridge entities with target entities and bridge entities, modeling various associations between them, and then use a graph recurrent network to encode this graph. Finally, we introduce a novel debiasing strategy to calibrate the original prediction distribution. Experimental results on the closed and open settings show that our model significantly outperforms all baselines, including the GPT-3.5-turbo and InstructUIE, achieving state-of-the-art performance. Particularly, our model obtains 66.23\% and 55.87\% AUC points in the official leaderboard\footnote{\url{https://codalab.lisn.upsaclay.fr/competitions/3770\#results}} under the two settings, respectively,
ranking the first place in all submissions since December 2023. Our code is available at \url{https://github.com/DeepLearnXMU/CoRE-NEPD}.

\end{abstract}

\section{Introduction}

Relation Extraction (RE) is a fundamental natural language processing (NLP) task, which aims to predict the relationship between two entities in a given context. Usually, conventional RE studies limit the context within a single sentence or document \citep{zengdao-2014,santos-2015,cai-2016,zhang-2018,wang-2022}. However, since a large number of relational facts are not described in the same document \citep{yao-2021}, many researchers have begun to concentrate on cross-document RE (CoRE), where the given target entities (\emph{head entity} and \emph{tail entity}) are located in different documents  ~\cite{yao-2021,wang-2022,Wu-2023}. 

In this regard, \citet{yao-2021} first explore this task. They not only release the dataset CodRED, where relevant documents of target entities connected by bridge entities are organized as text paths, but also propose BERT-based models for this task.  
However, their models suffer from the negative effect of irrelevant context in the model input and do not fully leverage the connections across text paths. To solve these issues, \citet{wang-2022} propose an Entity-based Cross-path Relation Inference Method (Ecrim). They employ an entity-centered noise filter to refine the model input and incorporate a cross-path entity relation attention to capture the connections across different text paths. Unlike the Ecrim,
\citet{Wu-2023} present a local-to-global causal reasoning model (LGCR),
which is a graph-based model using a local causality estimation algorithm to filter the noisy information.
Despite their success, their models still suffer from two issues.

\begin{figure*}
    \centering
\includegraphics[width=1.0\textwidth]{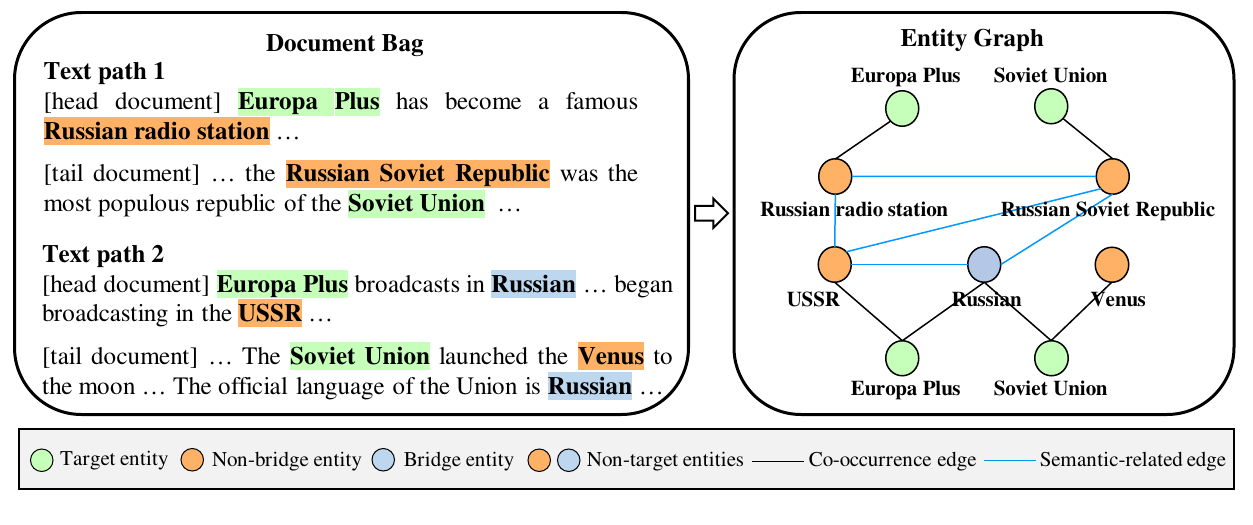}
    \vspace{-1cm}
    \caption{A document bag containing two text paths and its corresponding entity graph.
    Note that we use different nodes to represent the same target entity for different text paths, which facilitates learning the relation between target entities specific to the text path.
    }
    \label{img:task}
\end{figure*}

\textbf{First}, previous methods only consider the target entities and bridge entities, while ignoring the \emph{non-bridge entities} that solely co-occur with only one target entity. According to the statistics of the CodRED dataset \cite{yao-2021}, we observe that on average each text path contains 18.8 non-target entities: 2.6 bridge entities while 16.2 non-bridge entities that may also help the relation prediction. For example, in the text path 1 of Figure \ref{img:task}, the target entities ``\emph{Europa Plus}'' and ``\emph{Soviet Union}'' only occur with the non-bridge entities ``\emph{Russian radio station}'' and ``\emph{Russian Soviet Republic}'', respectively. Thus, it is difficult to correctly predict their relation due to the absence of bridge entities. If non-target entities ``\emph{Russian radio station}'' and ``\emph{Russian Soviet Republic}'' are semantically related, then there may also be some relationship between the target entities ``\emph{Europa Plus}'' and ``\emph{Soviet Union}''. Therefore, intuitively, such semantic correlation can be exploited to benefit the relation prediction between target entities. \textbf{Second}, 85\% of document bags are labeled as NA relation in the CodRED training set. As a result, the training dataset can only offer limited supervision signals for the model to learn non-NA relations, which leads to prediction bias.

To address these issues, in this paper, we propose a novel graph-based CoRE model with non-bridge entity enhancement and prediction debiasing. As illustrated in the right part of Figure \ref{img:task}, we represent each input document bag with a unified entity graph, where entity nodes are initialized with BERT representations. In this graph, each node indicates a target entity, bridge entity, or non-bridge entity, and two types of edges are introduced: 1) \emph{co-occurrence edges}, each of which connects a target entity and a non-target entity that co-occurs with it. 2) \emph{semantic-related edges}, which are used to connect any two semantic-related non-target entities. Then, we utilize Graph Recurrent Network (GRN) \cite{zhang-2018} to encode this graph, where the interdependency among connected nodes is captured based on the recurrent gating mechanism. Afterward, we aggregate the entity representations learned from GRN and utilize a cross-path entity relation attention module to capture the connections across text paths.

Besides, inspired by the studies on other NLP tasks \citep{utama-2020,xiong-2021,fei-2023},
we propose a simple yet effective prediction debiasing strategy to calibrate the relation prediction. 
Specifically, we introduce two prediction distributions: 1) $\overline{y}_{rela}$. To obtain this distribution, we fix the parameters of the original model and retrain a new classifier only using non-NA instances. Notably, the new classifier can avoid the negative impact of excessive NA instances and achieve better prediction performance on non-NA instances. 2) $\overline{y}_{bias}$. We mask the most important non-target entities of our entity graph as input to derive this distribution. Compared with the original prediction
distribution, this distribution is more biased and thus can be used for debiasing in a manner of subtraction.
Finally, we integrate the two newly-introduced prediction distributions to calibrate the original ones.
To the best of our knowledge, our work is the first attempt to study the prediction debiasing in this task. 

To investigate the effectiveness of our model, we conduct extensive experiments on both the closed and open settings of CodRED. Experimental results and in-depth analysis show that our model significantly outperforms all competitive baselines, including the LLMs. Particularly, compared with all submitted results since December 2023, our model ranks the first place in the official leaderboard.

\begin{figure*}[ht]
    \centering
    \includegraphics[width=1.0\textwidth]{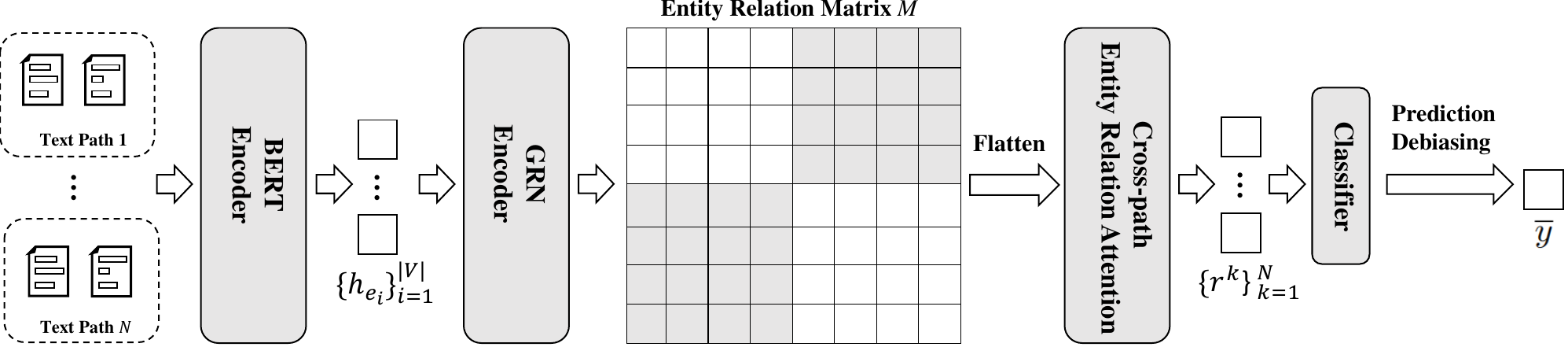}
    \caption{The overall architecture of our model.
$h_{e_i}$ is the entity representation after BERT encoding, $|V|$ is the number of entities nodes, $M$ is the entity relation matrix, $r^k$ is the final relation representation between target entities, $\overline{y}$ is the prediction distribution after debiasing
    }
    \label{img:method}
\end{figure*}

\section{Our Model}
\vspace{-0.1cm}
In this section, we introduce our proposed model in detail. As shown in Figure \ref{img:method}, our model contains three important components: a BERT encoder, a GRN encoder, and a classifier with a cross-path entity relation attention module, and then we detail the prediction debiasing strategy used in inference.

\subsection{BERT Encoder}
Given a target entity pair, we first obtain relevant documents and conduct data preprocessing\footnote{The detailed data preprocessing we used can be found in Appendix \ref{A.1}} to filter the noisy information over multiple documents.
As a common practice, we use BERT\footnote{We analyze the effect of the BERT version in Appendix \ref{A.6}} to learn the representations of entities, which will provide useful initial information for the subsequent GRN encoding.
Concretely, for each entity $e_i$, we first insert a special symbol ``*'' at the start and end of entity mentions and obtain the mention representations by max-pooling operation.
Then we collect all the mentions in a text path and follow \citet{Robin-2019} to obtain its path-level representation $h_{e_i}$, using $h_{e_i} = {\rm log}\sum_{j=1}^{N_{e_i}}
{\rm exp}\left(h_{m^{i}_{j}}\right)$, where $N_{e_i}$ denotes the mention number of $e_i$ in the text path, and $m^i_j$ is the representation of its $j$-th entity mention.
\subsection{GRN Encoder}
To construct our GRN encoder, we first represent the whole input document bag as a unified entity graph, which facilitates the introduction of non-bridge entities to strengthen the semantic association between target entities. Then, we introduce a GRN to encode the graph, where the learned entity representations will provide information for the subsequent relation prediction.

\subsubsection{Entity Graph}
To facilitate the subsequent description, we take the document bag shown in Figure \ref{img:task} as an example and describe how to use a unified entity graph to represent an undirected one $G$$=$$(V, E)$.

In the node set $V$, each node is a target entity, a bridge entity, or a non-bridge entity. Apparently, our entity graph considers more information than previous studies. Let us revisit the example in Figure \ref{img:task}, 
where we first identify the bridge entity ``\emph{Russian}”, the non-bridge entities ``\emph{Russian radio station}” and ``\emph{Russian Soviet Republic}”, and then include them with the target entities ``\emph{Europa Plus}” and ``\emph{Soviet Union}” into the entity graph. 
Note that we use different nodes to represent the same head entity for different text paths, which facilitates learning the relation between target entities specific to the text path.

To capture various kinds of interdependences between entities for cross-document RE, we consider two kinds of edges in the edge set $E$:
(1) each target entity and each non-target entity (bridge or non-bridge entities) within the same document are connected via a \emph{co-occurrence edge}; (2) any two semantic-related non-target entities are connected by a \emph{semantic-related edge}. During this process, we calculate the cosine similarity between the representations of any two non-target entities, and determine that they are semantically related if their similarity is greater than a threshold $\eta$. Back to Figure \ref{img:task}, ``\emph{Europa Plus}'' and ``\emph{Russian radio station}'' are connected by a co-occurrence edge, and ``\emph{Russian radio station}'' and ``\emph{Russian Soviet Republic}'' are connected by a semantic-related edge.

\subsubsection{Encoding with GRN}
We then introduce GRN \cite{zhang-2018} to encode this graph. Typically, it updates node representations using recurrent gating mechanisms, and thus has been widely used in many NLP tasks \citep{yongjing-2019,yongjing-2020,shaopeng-2021}.
Here, we choose GRU to update node representations since it has fewer parameters and better efficiency.

\begin{figure}
    \includegraphics[width=0.45\textwidth,height=0.3\textwidth]{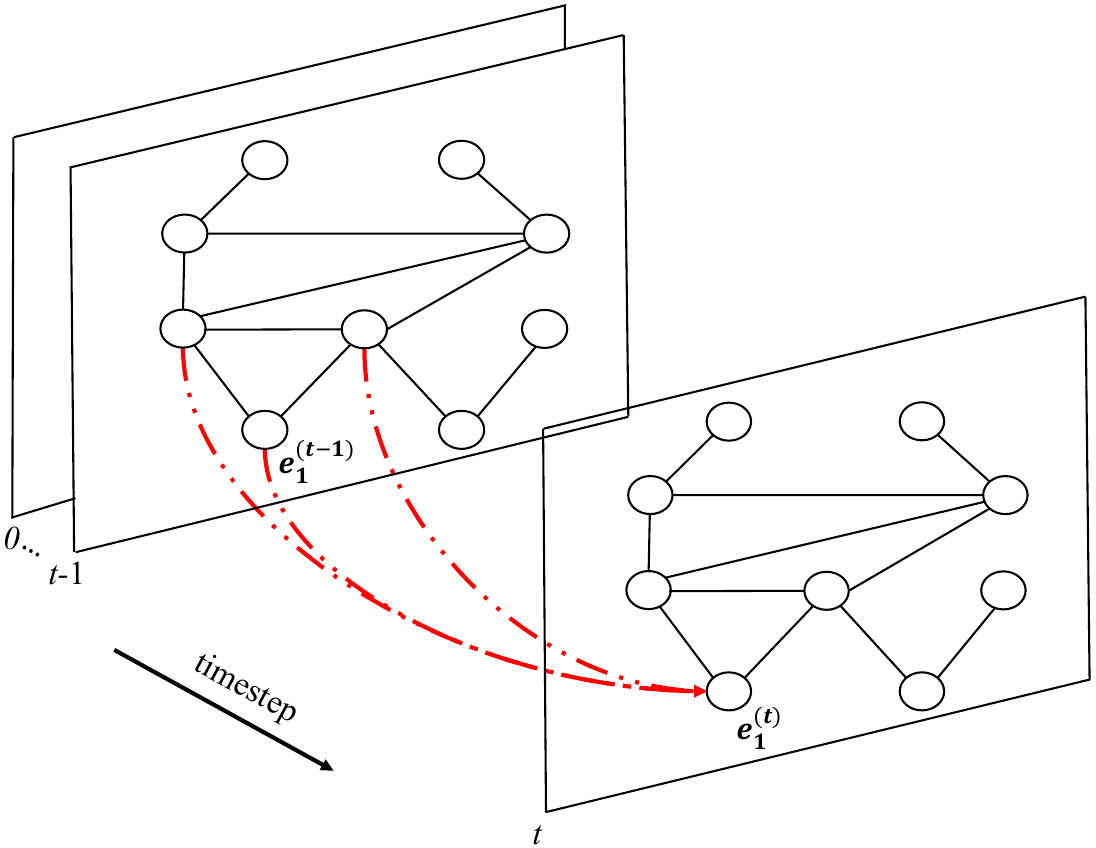}
    \vspace{-0.2cm}
    \caption{An example of message-passing procedures of our GRN encoder at the $t$-th timestep. 
    The update of $\bm{e}_{1}^{(t)}$ is in the red dash lines.}
    \label{img:tth_graph_update}
    \vspace{-1em}
\end{figure}

In Figure \ref{img:tth_graph_update}, we show the procedure of updating node representations in our graph at the $t$-th timestep. Specifically, for the node of entity $e_i$, we first gather the information from its connected entity nodes of the graph:
{
\setlength\abovedisplayskip{5pt}
\setlength\belowdisplayskip{0pt}
\begin{equation}
\bm{c}_{i}^{(t)} =\sum_{j \in A(e_i)} \bm{e}_{j}^{(t-1)},
\label{graph_update2}
\end{equation}
}\noindent where $\bm{c}_{i}^{(t)}$ is the collected context information used to update the node representation of entity $e_i$, $A(e_i)$ represents the set of neighboring nodes of entity $e_i$, $\bm{e}_{j}^{(t-1)}$ is the node representation of entity $e_j$ at the $(t\mspace{-4mu}-\mspace{-4mu}1)$-th timestep.
Besides, we initialize the entity nodes with their representations learned from BERT encoder: $\bm{e}^{(0)}_i$$=$$h_{e_i}$, where $1\mspace{-4mu}\leq\mspace{-4mu}{i}\mspace{-4mu}\leq\mspace{-4mu}{\left|V\right|}$.
Afterward, we update the node representation of entity $e_i$ in the following way:
\setlength\abovedisplayskip{3pt}
\setlength\belowdisplayskip{3pt}
\begin{align}
\bm{r}_{i}^{(t)} &= \sigma(W^r \bm{c}_{i}^{(t)} + U^r \bm{e}_{i}^{(t-1)}),
\label{ri_graph_update} \\
\bm{z}_{i}^{(t)} &= \sigma(W^z \bm{c}_{i}^{(t)} + U^z \bm{e}_{i}^{(t-1)}),
\label{zi_graph_update} \\
\bm{u}_{i}^{(t)} &= \tanh(W^u \bm{c}_{i}^{(t)} + U^u(\bm{r}_{i}^{(t)}
 \odot \bm{e}_{i}^{(t-1)})),
\label{ui_graph_update} \\
\bm{e}_{i}^{(t)} &= (1 - \bm{z}_{i}^{(t)}) \odot \bm{u}_{i}^{(t)} + \bm{z}_{i}^{(t)} \odot \bm{e}_{i}^{(t-1)},
\label{eit_graph_update}
\end{align}
where $\bm{r}_{i}^{(t)}$ and $\bm{z}_{i}^{(t)}$ are reset and update gates, $\odot$ is the Hadamard product, $\bm{u}_{i}^{(t)}$ denotes the temporary node representations of $\bm{e}_{i}$ at \emph{t}-th step, $W^*$  and $U^*$ are trainable parameter matrixes.

\subsection{Classifier with Cross-path Entity Relation Attention}
Following \citet{wang-2022}, on the basis of the entity representations learned from GRN, we utilize a cross-path entity relation attention to model the connections across different text paths, aggregating better relation representations of the target entity pair. Then, we feed these aggregated representations into an MLP classifier for prediction. 

Specifically, we first collect all the entity representations learned from the GRN encoder and then construct a entity relation matrix $M\mspace{-4mu}\in\mspace{-4mu}\mathbb{R}^{\left|V\right|\times\left|V\right|}$, with each element $r_{i,j}$ denoting the representation of the relation between the entities $e_i$ and $e_j$:
\setlength\abovedisplayskip{10pt}
\setlength\belowdisplayskip{10pt}
\begin{equation}
r_{i,j} = {\rm ReLU}(W({\rm ReLU}(W^u \bm{e}_i + W^v \bm{e}_j)),
\label{euv}
\end{equation}
where $\bm{e}_i$, $\bm{e}_j$ are the representations of entity $e_i$, $e_j$ respectively, $W^*$ are trainable parameter matrixes.

Then we flatten the entity relation matrix $M$ and perform self-attention on it, to capture the intra- and inter-path dependencies. By doing so, we can obtain the representation $r^k$ of the relation between the target entities of each text path $k$. Subsequently, we feed $r^k$ into the MLP classifier for prediction:
\begin{equation}
\overline{y}^k = {\rm MLP}(r^k),
\label{yi}
\end{equation}
where $\overline{y}^k$ is the predicted relation distribution for text path $k$.

Finally, to obtain the bag-level prediction $\overline{y}$, we follow \citet{wang-2022} to perform a max-pooling operation as follows:
\begin{equation}
\overline{y} = {\rm Max}(\{\overline{y}^k\}_{k=1}^N),
\label{yb}
\end{equation}
where $N$ denotes the number of text paths in a document bag.
\subsection{Model Inference with Prediction Debiasing}
Following common practice, we train our model by minimizing the cross-entropy loss of training data. However, as previous analysis, the training data contains numerous target entity pairs with NA relation, which leads to the prediction biasing during inference. To address this issue, we propose a simple yet effective debiasing strategy that introduces two prediction distributions to calibrate the original prediction ones in the following way:
\begin{equation}
\overline{y} = \overline{y} + \lambda\;({\overline{y}_{rela} - \overline{y}_{bias}}),
\label{unbias_method}
\end{equation}
where $\lambda$ is a hyper-parameter controlling the effects of newly-introduced prediction distributions.
Here, we give detailed descriptions to $\overline{y}_{rela}$ and $\overline{y}_{bias}$:
\begin{itemize}

    \item \textbf{$\overline{y}_{rela}$}. This prediction distribution is used to refine the prediction of the model on the instances with non-NA relations. 
    To obtain this distribution, we stack a new classifier on the original model. During the training process, we fix all parameters of the original model and only tune this classifier using the non-NA instances. Notably, unlike the original classifier, the training of this classifier avoids the negative impact of excessive NA instances. Thus, it can achieve better prediction performance on non-NA instances.
    \item \textbf{$\overline{y}_{bias}$}. Compared with the original prediction distribution,
this distribution is more biased and thus can be used to debias the original prediction distribution in a manner of subtraction. Notably, unlike the above $\overline{y}_{rela}$, we do not retrain the model or classifier to obtain $\overline{y}_{bias}$. To obtain this distribution, we first quantify the importance of each non-target entity with the average weight of target entities attending to the non-target entity.
Subsequently, we mask the most important 50\% non-target entities and then feed the remaining sub-graph into the GRN to derive $\overline{y}_{bias}$.
Apparently, due to this sub-graph lacking some important non-target entities, $\overline{y}_{bias}$ prefers NA relation and thus is more biased than $\overline{y}$.
\end{itemize}
\vspace{-0.3cm}

\section{Experiments} \label{exper}
\begin{table}
\centering
\resizebox{0.5\textwidth}{!}{
\begin{tabular}{l|c|c|c|c|c}
\toprule[1pt]
\multirow{1}{*}{}
& \multicolumn{3}{c}{\textbf{Closed}}
& \multicolumn{2}{|c}{\textbf{Open}}
\\
\cmidrule(r){2-5} \cmidrule(l){5-6}
\textbf{} & \textbf{Train} & \textbf{Dev} & \textbf{Test} & \textbf{Dev} & \textbf{Test}\\
\Xcline{1-6}{0.4pt}
{\#Bags(non-NA)} & {2,733} & {1,010} &{1,012} &{1,010} & \multirow{2}{*}{5,523}\\
\Xcline{1-5}{0.4pt}
{\#Bags(NA)} & {16,668} & {4,558} & {4,523} &{4,558}\\
\Xcline{1-6}{0.4pt}
{\#Text paths(non-NA)} & {8,263} & {2,558} & \multirow{2}{*}{40,524}& {15,072} & \multirow{2}{*}{7,7840}\\
\Xcline{1-3}{0.4pt} \Xcline{5-5}{0.4pt}
{\#Text paths(NA)} & {120,925} & {38,182} & {}&{62,863}\\
\Xcline{1-6}{0.4pt}
{\#Tokens/Doc} & {4,938.6} & {5,031.6} & {5,129.2} &{5,934.4} &{5,983.4} \\
\Xcline{1-6}{0.4pt}
{\#Path/Bag} & {6.67} & {7.31} & {7.32} &{13.99} &{14.09}\\
\toprule[1pt]
\end{tabular}}
\caption{Statistics of CodRED. Note that, Bags(non-NA) denotes the bags with non-NA relations, and Bags(NA) denotes the bags with NA relations.}
\label{tab:codred}
\end{table}
\subsection{Setup}
\textbf{Dataset.} To evaluate our model, we use the commonly-used dataset--CodRED \cite{yao-2021} to conduct experiments under two settings: 1) \textbf{closed setting}, where the related documents of target entities are given in advance for constructing the text path. 2) \textbf{open setting}, where we have to first retrieve related documents from Wikipedia and then evaluate the model performance. Table \ref{tab:codred} shows the detailed statistics of CodRED.
Note that the training data of CodRED contains 16,668 document bags with NA relation and 2,733 document bags with non-NA relations, where the significant number difference between NA instances and non-NA instances leads to the prediction bias.


\textbf{Settings.} When constructing our model, we set the similarity threshold $\eta$ for semantic-related edges to $0.6$.\footnote{We analyze the effect of $\eta$  in Appendix \ref{A.2}} As for the debiasing strategy, we set the mask rate of important non-target entities to 0.5 and $\lambda$ to 0.1, which will be analyzed in Section \ref{section:3.3}. Besides, we employ a 3-layer GRN to encode our unified entity graph and a 2-layer Transformer for cross-path entity relation attention, where the embedding size and hidden state dimension are both set to 768. To effectively train our model, we employ AdamW with a learning rate of 3e-5. To ensure a fair comparison, we follow \citet{yao-2021} to train the model on the closed setting, where extra document-level data are involved.\footnote{We investigate the effect of extra document-level data in Section \ref{A.3}}

As implemented in previous studies \citep{yao-2021, wang-2022}, we use four evaluation metrics for the development set: F1, AUC, P@500, and P@1000, and two evaluation metrics for the test set: F1 and AUC. Additionally, we compare our model with the LLMs using micro-F1.
Finally, following \citet{yao-2021},
we obtain the evaluation scores on the test set by submitting prediction results into Codalab.\footnote{\url{https://codalab.lisn.upsaclay.fr/competitions/3770}}




\begin{table*}[ht]
\centering
\resizebox{1\textwidth}{!}{
\normalsize
\begin{tabular}{@{}lcccccccccccc@{}}
\toprule[1pt]

{\multirow{3}{*}{$\quad$Model}}
& \multicolumn{6}{c}{\textbf{Closed }}         
& \multicolumn{6}{c}{\textbf{Open $\quad$}}    
\\ \cmidrule(lr){2-7} \cmidrule(l){8-13}

& \multicolumn{4}{c}{\textbf{Dev}}  
& \multicolumn{2}{c}{\textbf{Test}} 

& \multicolumn{4}{c}{\textbf{Dev}}   & \multicolumn{2}{c}{\textbf{Test$\quad$}}   \\
\cmidrule(lr){2-5} \cmidrule(lr){6-7} \cmidrule(lr){8-11} \cmidrule(l){12-13}  

\multicolumn{1}{r}{}                       
& \textbf{F1} & \textbf{AUC} & \textbf{P@500} & \textbf{P@1000} & \textbf{F1}         & \textbf{AUC}         & \textbf{F1} & \textbf{AUC} & \textbf{P@500} & \textbf{P@1000} & \textbf{F1} & \textbf{AUC$\quad$}                  
\\ \midrule
                        {$\quad$End-to-end  \cite{yao-2021}}$\dagger$ & {51.26} & {47.94} & {62.80} & {51.00} & {51.02} & {47.46}
                        & {47.23} & {40.86} & {59.00} & {46.30} & {45.06} & {39.05$\quad$}                     
                        \\
 {$\quad$Ecrim  \cite{wang-2022}}$\dagger$ & {61.12} & {60.91} & {\textbf{78.89}} & {60.17} & {62.48} & {60.67} 
& {\textendash} & {\textendash} & {\textendash} & {\textendash} & {\textendash} & {\textendash}\\
{$\quad$Ecrim  \cite{wang-2022}} & {61.42} & {61.05} & {78.04} & {60.44} & {62.73} & {60.84} 
& {51.28} & {49.65} & {69.25} & {51.55} & {51.78} & {49.58$\quad$}  

\\
 {$\quad$LGCR  \cite{Wu-2023}}$\dagger$ & {61.67} & {63.17} & {76.65} & {61.84} & {61.08} & {60.75}& {52.96} & {51.48} & {\textbf{70.06}} & {52.19} & {53.45} & {50.15$\quad$} \\

{$\quad$LGCR  \cite{Wu-2023}} & {61.72} & {63.05} & {76.83} & {61.97} & {61.25} & {60.44}& {53.02} & {51.26} & {69.67} & {52.25} & {53.60} & {50.12$\quad$} 
\\
       \midrule{$\quad$Ours} & {\textbf{63.63}} & {\textbf{65.01}}  & {77.84} & {\textbf{64.03}} & {\textbf{64.41}} & {\textbf{66.23}} 
& {\textbf{54.49}} & {\textbf{54.92}}  & {68.66} & {\textbf{53.84}} & {\textbf{56.68}} & {\textbf{55.87$\quad$}}   
\\ 
\toprule[1pt]
\end{tabular}}
\caption{Experimental results on the  CodRED dataset. $\dagger$ indicates previously reported scores.}
\label{tab:main_result}
\end{table*}

\subsection{Baselines}
We compare our model with the following baselines:


\begin{itemize}[itemsep=2pt,topsep=0pt,parsep=0pt]
    \item \textbf{End-to-end} \citep{yao-2021}. This model employs BERT to obtain the representations of text paths. Then, a selective attention module is used to obtain the aggregated representations of target entities. Finally, the aggregated representations are fed into a classifier to predict the relation between target entities. 
    \item \textbf{Ecrim} \citep{wang-2022}. It is our most important baseline. This model first uses an entity-based document-context filter to retain useful information in the given documents by using the bridge entities in the text paths. Then, it is equipped with a cross-path entity relation attention, which allows entity relations across text paths to interact with each other. 
    \item \textbf{LGCR} \citep{Wu-2023}. This model first estimates the causal effect for each semantic unit (text path, head entity, tail entity, and bridge entity). Then, it constructs a global reasoning graph based on the co-occurrence of entity mentions and the structure of text paths, and uses the relative causal association calculated by local causal effect to control the message propagation ability between nodes.  
\end{itemize}
The number of trainable parameters for our model and the above baselines are 
$1.30 \times 10^8$, $1.08 \times 10^8$, $1.23 \times 10^8$, and $1.19 \times 10^8$, respectively.

To further verify the performance of our model, we compare it with powerful LLMs. In addition to the commonly-used \textbf{GPT-3.5-turbo} \cite{ouyang-2022}, we consider a variant of InstructUIE \citep{wang-2023}, termed as \textbf{InstructUIE-FT}, which is fine-tuned on CodRED to enhance its cross-document RE ability.
Note that InstructUIE is a strong information extraction framework that captures the inter-task dependency to uniformly model various information extraction tasks. 

To investigate the few-shot ability of LLMs in classification tasks, the standard practice based on LLMs introduces an exemplar for each candidate label to investigate the few-shot ability of LLMs \cite{yoo-2022,dong-2023,fan-2023}. However, in the CoRE task, the number of candidate labels is 276, making it impractical to introduce an exemplar for each label due to the maximum context length of the LLMs. Therefore, we only test the zero-shot performance of LLMs. The detailed settings for these LLMs are shown in Appendix \ref{A.5}.

\begin{figure}[t]
    \vspace{-0.1cm}
    \centering\includegraphics[width=0.45\textwidth,height=0.25\textwidth]{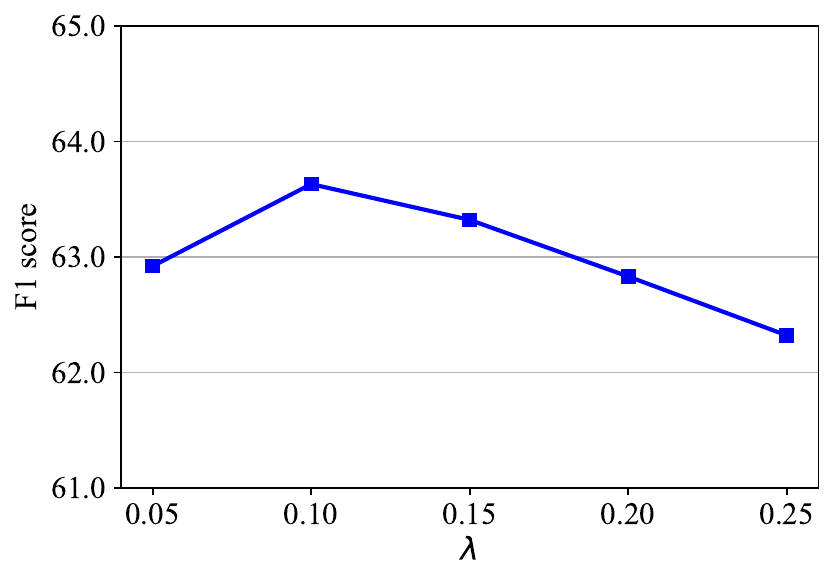}
    \vspace{-0.2cm}
    \caption{The performance of our model (F1 score) on the CodRED development set under the closed setting, using different $\lambda$ .}
    \label{img:lambda}
\end{figure}

\subsection{Effect of the Adaptive Parameter $\lambda$}
\label{section:3.3}
We first investigate the impact of the hyper-parameter
$\lambda$ (See Equation \ref{unbias_method}) on the development set under the closed setting. To this end, we gradually vary $\lambda$ from 0.05 to 0.25 with an increment of 0.05 in each step. 

As shown in Figure \ref{img:lambda}, we find that our model achieves the best performance when $\lambda$=0.1. Therefore, we set $\lambda$=0.1 for all experiments thereafter.

\begin{table}[t]
\centering
\resizebox{.4\textwidth}{!}{
\normalsize
\begin{tabular}{lccr}
\toprule[1pt]
    \multirow{2}{*}{Model} & 
    \multicolumn{1}{c}{\textbf{Closed-Dev}} & 
    \multicolumn{1}{c}{\textbf{Open-Dev}} & \\
    \cline{2-4} &
    \textbf{micro-F1} & \textbf{micro-F1}& \\
\hline
{End-to-end} & {78.24} & {77.63}\\
{Ecrim} & {82.59} & {81.08}\\

{LGCR} & {82.07} & {81.79}\\

\hline
{GPT-3.5-turbo} & {28.05} & {25.31}\\
{InstructUIE} & {70.34} & {64.05}\\
{InstructUIE-FT} & {80.72} & {79.66}\\
\hline
{Ours} & {\textbf{84.35}} & {\textbf{83.92}}\\
\toprule[1pt]
\end{tabular}}
\caption{Experimental results with LLMs.}
\label{tab:llm}
\end{table}

\begin{table*}[ht]
\centering
\resizebox{1\textwidth}{!}{
\normalsize
\begin{tabular}{@{}lcccccccccccc@{}}
\toprule[1pt]
{\multirow{3}{*}{$\quad$ Model}}
& \multicolumn{6}{c}{\textbf{Closed }}          
& \multicolumn{6}{c}{\textbf{Open }}    \\ 
\cmidrule(lr){2-7} \cmidrule(l){8-13}    

& \multicolumn{4}{c}{\textbf{Dev}}  
& \multicolumn{2}{c}{\textbf{Test}} 

& \multicolumn{4}{c}{\textbf{Dev}}   & \multicolumn{2}{c}{\textbf{Test $\quad$ }}   \\
\cmidrule(lr){2-5} \cmidrule(lr){6-7} \cmidrule(lr){8-11} \cmidrule(l){12-13}  
                     
& \textbf{F1} & \textbf{AUC} & \textbf{P@500} & \textbf{P@1000} & \textbf{F1}         & \textbf{AUC}         & \textbf{F1} & \textbf{AUC} & \textbf{P@500} & \textbf{P@1000} & \textbf{F1} & \textbf{AUC$\quad$ }                  
\\ \midrule
{$\quad$ Ours} & {63.63} & {65.01}  & {77.84} & {64.03} & {64.41} & {66.23}
& {54.49} & {54.92}  & {68.66} & {53.84} & {56.68} & {55.87$\quad$ } \\
\hline
 {$\quad$ $\quad$ w/o $\overline{y}_{rela}$} & {63.12} & {63.77}  & {77.54} & {63.43} & {63.30} & {64.69} 
& {53.74} & {53.31}  & {67.06} & {52.94} & {55.01} & {54.95$\quad$ }               
\\
{$\quad$ $\quad$ w/o $\overline{y}_{bias}$} & {63.32} & {64.85}  & {77.44} & {63.63} & {63.85} & {65.78} & {54.07} & {54.16}  & {67.66} & {53.04} & {55.66} & {55.38$\quad$ } \\
{$\quad$ $\quad$ w/o NBE} & {63.03} & {64.36}  & {77.42} & {63.31} & {63.81} & {65.28} & {53.85} & {54.02}  & {67.24} & {52.87} & {55.51} & {55.14$\quad$ } \\
{$\quad$ $\quad$ w/o GRN} & {62.88} & {63.79}  & {77.35} & {62.89} & {63.15} & {64.87} & {53.33} & {53.10}  & {67.02} & {52.53} & {54.89} & {54.44$\quad$ } \\
{$\quad$ $\quad$ w/o $\overline{y}_{rela}$, $\overline{y}_{bias}$} & {61.88} & {63.12}  & {77.40} & {62.23} & {62.23} & {62.77}& {52.26} & {52.45}  & {66.86} & {52.14} & {53.25} & {52.85$\quad$ }

\\
{$\quad$ $\quad$ w/o $\overline{y}_{rela}$, $\overline{y}_{bias}$, NBE} & {61.15} & {62.36} & {77.04} & {61.42} & {61.62} & {61.87} & {51.49} & {51.14} & {66.07} & {51.45} & {52.12} & {52.08$\quad$ }\\
{$\quad$ $\quad$ w/o $\overline{y}_{rela}$, $\overline{y}_{bias}$, SRE} & {61.27} & {62.85} & {77.15} & {61.68} & {62.01} & {62.29}& {51.82} & {51.95} & {66.49} & {51.76} & {52.88} & {52.32$\quad$ }\\
    \toprule[1pt]
\end{tabular}}
\caption{Ablation study of our model on the CodRED dataset. Note that \textbf{NBE} refers to the non-bridge entities and \textbf{SRE} refers to the semantic-related edges in the graph.}
\label{tab:ablation_study}
\end{table*}
\subsection{Main Results}
Table \ref{tab:main_result} shows the experimental results under two settings. 
Note that the performance of our reproduced LGCR and Ecrim model are comparable to that of their original paper,
proving that our experimental comparison is convincing.
Overall, under both settings,
our model exhibits better performance in most metrics than all baselines, except for P@500. Most importantly, we submit the test results to the official competition leaderboard,
where our model obtains 66.23\% and 55.87\% AUC points under the closed and open settings, respectively, ranking the first place in all submitted results since December 2023. 


Moreover,
we compare our model with LLMs in Table \ref{tab:llm}.
Under both settings,
we can find that our model significantly outperforms the LLMs,  including InstructUIE-FT that has also been fine-tuned on CodRED. These results once again demonstrate the effectiveness of our model.

\begin{table*}[ht]
\centering
\resizebox{.8\textwidth}{!}{
\normalsize
\begin{tabular}{@{}lcccccccc@{}}
\toprule[1pt]

{\multirow{3}{*}{$\quad$Model}}
& \multicolumn{4}{c}{\textbf{Closed-set}}         
& \multicolumn{4}{c}{\textbf{Open-set$\quad$}}    
\\ \cmidrule(lr){2-5} \cmidrule(l){6-9}

& \multicolumn{2}{c}{\textbf{Dev}}  
& \multicolumn{2}{c}{\textbf{Test}} 

& \multicolumn{2}{c}{\textbf{Dev}}   & \multicolumn{2}{c}{\textbf{Test$\quad$}}   \\
\cmidrule(lr){2-3} \cmidrule(lr){4-5} \cmidrule(lr){6-7} \cmidrule(l){8-9}  

\multicolumn{1}{r}{}                       
& \textbf{F1} & \textbf{AUC} & \textbf{F1}         & \textbf{AUC}         & \textbf{F1} & \textbf{AUC}  & \textbf{F1} & \textbf{AUC$\quad$}                  
\\ \midrule
{$\quad$End-to-end  \cite{yao-2021}$\dagger$} & {26.56} & {15.67}& {\textendash} & {\textendash}
& {22.06} & {11.43}  & {\textendash} & {\textendash$\quad$}                     
                        \\
 {$\quad$Ecrim  \cite{wang-2022}} & {39.19} & {29.85}& {36.41} & {27.40} 
& {25.01} & {18.04} & {24.93} & {18.97$\quad$}                    
                                  \\{$\quad$LGCR  \cite{Wu-2023}} & {40.73} & {32.81} & {36.67} & {28.01}& {27.54} & {23.57} &  {26.85} & {23.32$\quad$} \\
                       \midrule{$\quad$Ours} & {\textbf{42.26}} & {\textbf{34.13}}  &  {\textbf{38.22}} & {\textbf{33.15}} 
            & {\textbf{30.26}} & {\textbf{25.54}}  & {\textbf{29.05}} & {\textbf{26.64$\quad$}}                 \\ \toprule[1pt]
\end{tabular}}
\caption{Experimental results with cross-document-only supervision on the  Codred dataset. $\dagger$ indicates previously reported scores.}
\label{tab:crossonly}
\end{table*}

\begin{figure}[t]
    \includegraphics[width=0.45\textwidth]
    {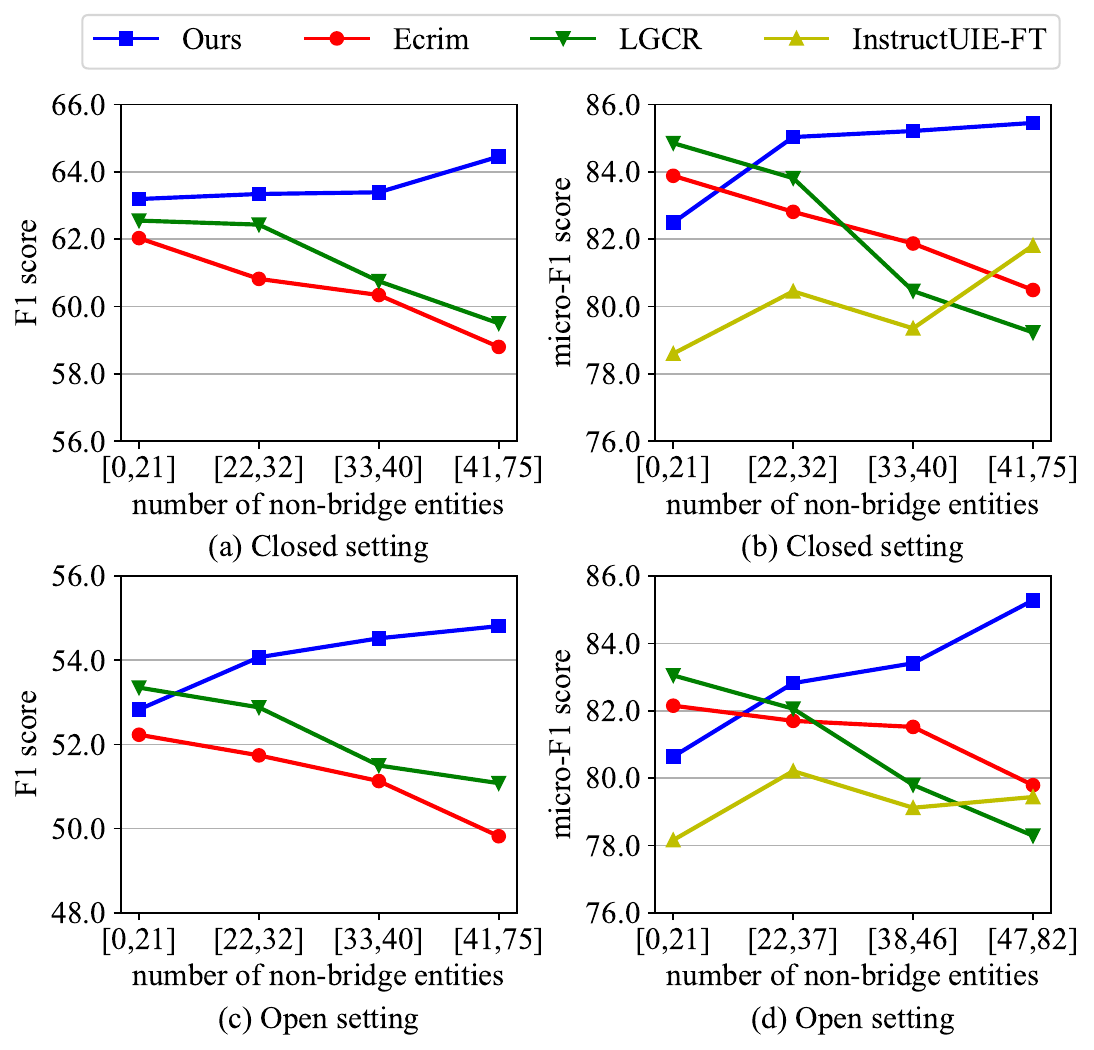}
    \caption{The model performance on different development subsets of CodRED under two settings.}
    \label{img:devf1subset}
\end{figure}

\subsection{Impact of Non-bridge Entity Number}
To assess the impact of non-bridge entities on our model, we first sort the development set according to the number of non-bridge entities in ascending order, and then equally split it into four subsets. After that, we compare the model performance on these subsets. To clearly display the experimental results, we only compare ours with Ecrim, LGCR and InstructUIE-FT, which are competitive baselines according to the results reported in Table \ref{tab:llm}.

As shown in Figure \ref{img:devf1subset},
regardless of the setting, as the number of non-bridge entities increases, 
the performance advantage of our model becomes increasingly apparent.
These results strongly demonstrate the generality and effectiveness of our model.

\subsection{Ablation Study}
To investigate the effectiveness of different components of our model, we further compare our model with the following variants in Table \ref{tab:ablation_study}:

\begin{itemize}[itemsep=2pt,topsep=0pt,parsep=0pt]
\item \emph{w/o $\overline{y}_{rela}$.} In this variant, we only use $\overline{y}_{bias}$ for prediction debiasing. 
\item \emph{w/o $\overline{y}_{bias}$.} We only use $\overline{y}_{rela}$ for prediction debiasing in this variant.  
\item \emph{w/o $\overline{y}_{rela}$, $\overline{y}_{bias}$.} In this variant, we do not calibrate the model prediction. 
\item \emph{w/o NBE.} In this variant, we remove all non-bridge entities from our entity graph.
\item \emph{w/o GRN.} In this variant, we remove the GRN Encoder from our model.

\item \emph{w/o $\overline{y}_{rela}$, $\overline{y}_{bias}$, NBE(Non-bridge entities).} In this variant, we directly remove all non-bridge entities from our entity graph. In other words, we only consider target entities and bridge entities in this variant. Notice that the main difference between this variant and Ecrim is that it adds an additional GRN on the top of BERT encoder. 

\item \emph{w/o $\overline{y}_{rela}$, $\overline{y}_{bias}$, SRE(Semantic-related edges).} When constructing this variant, we remove the debiasing module and all semantic-related edges from our entity graph. 



\end{itemize}

Table \ref{tab:ablation_study} lists the experimental results, where,
all variants are inferior to our model, verifying the effectiveness of each proposed module.
\subsection{Effect of Extra Document-level Data} 
\label{A.3}
To investigate the effect of extra document-level data in training, we only use the cross-document data to train various models.

From Table \ref{tab:crossonly}, we observe that the performance of various models on the CoRE task sharply decreases, which demonstrates that the extra document-level data can help models to capture more useful information. Moreover, we can observe that our model still significantly outperforms all baselines in this setting, confirming the capability of our model.



\subsection{Case Study}
Table \ref{tab:case} in the Appendix displays the prediction results of different models on two text paths sampled from the test set. In text path 1, due to the lack of bridge entities, Ecrim is unable to connect the head and tail entities, leading to the incorrect relation prediction. By contrast, our model leverages substantial non-bridge entities of text path 1 to build connections between target entities. Thus, both Ours and w/o $\overline{y}_{rela}$,  $\overline{y}_{bias}$ successfully predict the relation of text path 1, demonstrating that non-bridge entities can indeed provide useful supplementary information for the cross-document RE. Meanwhile, in text path 2, both Ecrim and w/o $\overline{y}_{rela}$, $\overline{y}_{bias}$ mistakenly predict the relation between target entities as NA. Only when the prediction debiasing strategy is used, our model can calibrate the prediction, thereby correctly predicting the relation as continent. This is consistent with the results reported in Table \ref{tab:ablation_study}, further verifying the effectiveness of our strategy.

\section{Related Work}
Most studies on RE mainly focus on the sentence-level RE \citep{cai-2016,zhang-2018,Zhangzhen-2019, Yamada-2020,lzhang_emnlp-2023} and document-level RE \citep{zengshuang-2020,Xu-2020,lzhang_emnlp-2022,lzhang_aaai-2023}, 
which are committed to identifying the relation between target entities from a sentence and document, respectively. Unlike these studies,
in this work, we concentrate on cross-document RE that aims at predicting the relation between target entities located in different documents. \citet{yao-2021} comprehensively investigated this task and released the first human-annotated CoRE dataset, CodRED. Besides, they explore an end-to-end model that jointly considers documents in text paths to predict the relation. However, it not only suffers from the negative effect of irrelevant context in text paths but also does not fully leverage the interconnections across text paths. To address the above-mentioned issues, \citet{wang-2022} propose Entity-based Cross-path Relation InferenceMethod (Ecrim). Typically, it uses an attention mechanism to capture the connection among different text paths through bridge entities, which is important to predict the final relation. Recently, \citet{Wu-2023} put forward local-to-global causal reasoning (LGCR). 
To aggregate information over multiple text paths, they construct a global heterogeneous graph, where a local causality estimation algorithm is proposed to assess the importance of different nodes in the graph.

However, their methods suffer from two limitations: 1) ignore the non-bridge entities, which exist broadly in each text path and can offer semantic associations between target entities, especially in the absence of bridge entities. 2) ignore the bias caused by the prominent number difference between NA and non-NA instances. 
In this work, we propose a graph-based model to fully exploit non-bridge entities for cross-document RE.
Besides, along the research line of debiasing in NLP \cite{Minot-2021,xiong-2021,fei-2023},
we propose a simple yet effective prediction debiasing strategy to refine the prediction of our model.


\section{Conclusion and Future work}
In this paper, we propose a novel graph-based cross-document RE model. Concretely, we first represent the input bag as a unified entity graph, where abundant non-bridge entities are introduced to provide useful information. Then, we use GRN to encode this graph, where the learned entity representations form the basis for subsequent relationship prediction. Besides, we propose a simple yet effective debiasing strategy to refine the original prediction distribution. To the best of our knowledge, the graph-based model with non-bridge entities and our debiasing strategy has not been explored before. Extensive experiments on the commonly-used dataset CodRED demonstrate the superiority of our model. In the future, we will study how to introduce more external knowledge to refine our model. 

\section*{Limitations}
One limitation of the present work lies in that we only rely on the attention score to measure the importance of nodes in the graph, and do not consider dynamically adjusting the importance score. Additionally, in the entity-based graph,  the edges are connected using a heuristic method, which may overlook useful information.


\section*{Acknowledgements}
The project was supported by National Natural Science Foundation of China (No. 62276219), and the Public Technology Service Platform Project of Xiamen (No. 3502Z20231043).
We also thank the reviewers for their insightful comments.
\bibliography{anthology,custom}

\begin{thebibliography}{30}
\expandafter\ifx\csname natexlab\endcsname\relax\def\natexlab#1{#1}\fi

\bibitem[{Cai et~al.(2016)Cai, Zhang, and Wang}]{cai-2016}
Rui Cai, Xiaodong Zhang, and Houfeng Wang. 2016.
\newblock Bidirectional recurrent convolutional neural network for relation classification.
\newblock In \emph{ACL 2016}.

\bibitem[{Dong et~al.(2023)Dong, Li, Dai, Zheng, Wu, Chang, Sun, Xu, Li, and Sui}]{dong-2023}
Qingxiu Dong, Lei Li, Damai Dai, Ce~Zheng, Zhiyong Wu, Baobao Chang, Xu~Sun, Jingjing Xu, Lei Li, and Zhifang Sui. 2023.
\newblock A survey on in-context learning.
\newblock \emph{arXiv preprint arXiv:2301.00234}.

\bibitem[{Fan et~al.(2023)Fan, Tian, Li, He, and Jin}]{fan-2023}
Caoyun Fan, Jidong Tian, Yitian Li, Hao He, and Yaohui Jin. 2023.
\newblock Comparable demonstrations are important in in-context learning: A novel perspective on demonstration selection.
\newblock \emph{arXiv preprint arXiv:2312.07476}.

\bibitem[{Han et~al.(2023)Han, Peng, Yang, Wang, Liu, and Wan}]{han-2023}
Ridong Han, Tao Peng, Chaohao Yang, Benyou Wang, Lu~Liu, and Xiang Wan. 2023.
\newblock Is information extraction solved by chatgpt? an analysis of performance, evaluation criteria, robustness and errors.
\newblock \emph{arXiv preprint arXiv:2305.14450}.

\bibitem[{Hu et~al.(2021)Hu, Shen, Wallis, Allen-Zhu, Li, Wang, Wang, and Chen}]{hu-2021}
Edward~J. Hu, Yelong Shen, Phillip Wallis, Zeyuan Allen-Zhu, Yuanzhi Li, Shean Wang, Lu~Wang, and Weizhu Chen. 2021.
\newblock Lora: Low-rank adaptation of large language models.
\newblock \emph{arXiv preprint arXiv:2106.09685}.

\bibitem[{Ikuya et~al.(2020)Ikuya, Akari, Hiroyuki, Hideaki, and Yuji}]{Yamada-2020}
Yamada Ikuya, Asai Akari, Shindo Hiroyuki, Takeda Hideaki, and Matsumoto Yuji. 2020.
\newblock Luke: Deep contextualized entity representations with entityaware self-attention.
\newblock In \emph{EMNLP 2020}.

\bibitem[{Joshua~R et~al.(2021)Joshua~R, Nicholas, Marc, Danne~C, Christopher~M, and Peter~Sheridan}]{Minot-2021}
Minot Joshua~R, Cheney Nicholas, Maier Marc, Elbers Danne~C, Danforth Christopher~M, and Dodds Peter~Sheridan. 2021.
\newblock Interpretable bias mitigation for textual data: Reducing gender bias in patient notes while maintaining classification performance.
\newblock \emph{arXiv preprint arXiv:2103.05841}.

\bibitem[{Lai et~al.(2021)Lai, Wang, Meng, Zhou, Ge, Zeng, Yao, Huang, and Su}]{shaopeng-2021}
Shaopeng Lai, Ante Wang, Fandong Meng, Jie Zhou, Yubin Ge, Jiali Zeng, Junfeng Yao, Degen Huang, and Jinsong Su. 2021.
\newblock Improving graph-based sentence ordering with iteratively predicted pairwise orderings.
\newblock In \emph{EMNLP 2021}.

\bibitem[{Ouyang et~al.(2022)Ouyang, Wu, Jiang, Almeida, Wainwright, Mishkin, Zhang, Agarwal, Slama, Ray, Schulman, Hilton, Kelton, Miller, Simens, Askell, Welinder, Christiano, Leike, and Lowe}]{ouyang-2022}
Long Ouyang, Jeff Wu, Xu~Jiang, Diogo Almeida, Carroll~L. Wainwright, Pamela Mishkin, Chong Zhang, Sandhini Agarwal, Katarina Slama, Alex Ray, John Schulman, Jacob Hilton, Fraser Kelton, Luke Miller, Maddie Simens, Amanda Askell, Peter Welinder, Paul Christiano, Jan Leike, and Ryan Lowe. 2022.
\newblock Training language models to follow instructions with human feedback.
\newblock \emph{arXiv preprint arXiv:2203.02155}.

\bibitem[{Robin et~al.(2019)Robin, Cliff, and Hoifung}]{Robin-2019}
Jia Robin, Wong Cliff, and Poon Hoifung. 2019.
\newblock Document-level n-ary relation extraction with multiscale representation learning.
\newblock In \emph{NAACL 2019}.

\bibitem[{Santos et~al.(2015)Santos, Xiang, and Zhou}]{santos-2015}
Cicero Nogueira~dos Santos, Bing Xiang, and Bowen Zhou. 2015.
\newblock Classifying relations by ranking with convolutional neural networks.
\newblock In \emph{ACL 2015}.

\bibitem[{Utama et~al.(2020)Utama, Moosavi, and Gurevych}]{utama-2020}
Prasetya~Ajie Utama, Nafise~Sadat Moosavi, and Iryna Gurevych. 2020.
\newblock Towards debiasing {NLU} models from unknown biases.
\newblock In \emph{EMNLP 2020}.

\bibitem[{Wang et~al.(2023{\natexlab{a}})Wang, Huang, Yan, Zhou, and Chen}]{fei-2023}
Fei Wang, James~Y. Huang, Tianyi Yan, Wenxuan Zhou, and Muhao Chen. 2023{\natexlab{a}}.
\newblock Robust natural language understanding with residual attention debiasing.
\newblock In \emph{ACL 2023 findings}.

\bibitem[{Wang et~al.(2022)Wang, Li, Fei, Li, Wu, Su, Shi, Ji, and Cai}]{wang-2022}
Fengqi Wang, Fei Li, Hao Fei, Jingye Li, Shengqiong Wu, Fangfang Su, Wenxuan Shi, Donghong Ji, and Bo~Cai. 2022.
\newblock Entity-centered cross-document relation extraction.
\newblock In \emph{EMNLP 2022}.

\bibitem[{Wang et~al.(2023{\natexlab{b}})Wang, Zhou, Zu, Xia, Chen, Zhang, Zheng, Ye, Zhang, Gui, Kang, Yang, Li, and Du}]{wang-2023}
Xiao Wang, Weikang Zhou, Can Zu, Han Xia, Tianze Chen, Yuansen Zhang, Rui Zheng, Junjie Ye, Qi~Zhang, Tao Gui, Jihua Kang, Jingsheng Yang, Siyuan Li, and Chunsai Du. 2023{\natexlab{b}}.
\newblock Multi-task instruction tuning for unified information extraction.
\newblock \emph{arXiv preprint arXiv:2304.08085}.

\bibitem[{Wang et~al.(2021)Wang, Kehai, and Tiejun}]{Xu-2020}
Xu~Wang, Chen Kehai, and Zhao Tiejun. 2021.
\newblock Discriminative reasoning for document-level relation extraction.
\newblock In \emph{ACL 2020}.

\bibitem[{Wu et~al.(2023)Wu, Chen, Hu, Shi, Xu, and Xu}]{Wu-2023}
Haoran Wu, Xiuyi Chen, Zefa Hu, Jing Shi, Shuang Xu, and Bo~Xu. 2023.
\newblock Local-to-global causal reasoning for cross-document relation extraction.
\newblock \emph{IEEE/CAA JOURNAL OF AUTOMATICA SINICA}, 10:1608--1621.

\bibitem[{Xiong et~al.(2021)Xiong, Chen, Pang, Cheng, Ma, and Lan}]{xiong-2021}
Ruibin Xiong, Yimeng Chen, Liang Pang, Xueqi Cheng, ZhiMing Ma, and Yanyan Lan. 2021.
\newblock Uncertainty calibration for ensemble-based debiasing methods.
\newblock In \emph{NIPS 2021}.

\bibitem[{Yao et~al.(2021)Yao, Du, Lin, Li, Liu, Zhou, and Sun}]{yao-2021}
Yuan Yao, Jiaju Du, Yankai Lin, Peng Li, Zhiyuan Liu, Jie Zhou, and Maosong Sun. 2021.
\newblock Codred: A cross-document relation extraction dataset for acquiring knowledge in the wild.
\newblock In \emph{EMNLP 2021}.

\bibitem[{Yin et~al.(2020)Yin, Lai, Song, Zhou, Han, Yao, and Su}]{yongjing-2020}
Yongjing Yin, Shaopeng Lai, Linfeng Song, Chulun Zhou, Xianpei Han, Junfeng Yao, and Jinsong Su. 2020.
\newblock An external knowledge enhanced graph-based neural network for sentence ordering.
\newblock \emph{Journal of Artificial Intelligence Research}.

\bibitem[{Yin et~al.(2019)Yin, Song, Su, Zeng, Zhou, and Luo}]{yongjing-2019}
Yongjing Yin, Linfeng Song, Jinsong Su, Jiali Zeng, Chulun Zhou, and Jiebo Luo. 2019.
\newblock Graph-based neural sentence ordering.
\newblock In \emph{IJCAI 2019}.

\bibitem[{Yoo et~al.(2022)Yoo, Kim, Kim, Cho, Jo, Lee, Lee, and Kim}]{yoo-2022}
Kang~Min Yoo, Junyeob Kim, Hyuhng~Joon Kim, Hyunsoo Cho, wiyeol Jo, Sang-Woo Lee, Sang-goo Lee, and Taeuk Kim. 2022.
\newblock Ground-truth labels matter: A deeper look into input-label demonstrations.
\newblock In \emph{EMNLP 2022}.

\bibitem[{Yuan et~al.(2023)Yuan, Xie, and Ananiadou}]{yuan-2023}
Chenhan Yuan, Qianqian Xie, and Sophia Ananiadou. 2023.
\newblock Zero-shot temporal relation extraction with chatgpt.
\newblock \emph{arXiv preprint arXiv:2304.05454}.

\bibitem[{Zeng et~al.(2014)Zeng, Liu, Lai, Zhou, and Zhao}]{zengdao-2014}
Daojian Zeng, Kang Liu, Siwei Lai, Guangyou Zhou, and Jun Zhao. 2014.
\newblock relation classification via convolutional deep neural network.
\newblock In \emph{ACL 2014}.

\bibitem[{Zeng et~al.(2020)Zeng, Xu, Chang, and Li}]{zengshuang-2020}
Shuang Zeng, Runxin Xu, Baobao Chang, and Lei Li. 2020.
\newblock Double graph based reasoning for document-level relation extraction.
\newblock In \emph{EMNLP 2020}.

\bibitem[{Zhang et~al.(2022)Zhang, Su, Chen, Miao, Min, Hu, and Shi}]{lzhang_emnlp-2022}
Liang Zhang, Jinsong Su, Yidong Chen, Zhongjian Miao, Zijun Min, Qingguo Hu, and Xiaodong Shi. 2022.
\newblock Towards better document-level relation extraction via iterative inference.
\newblock In \emph{EMNLP 2022}.

\bibitem[{Zhang et~al.(2023{\natexlab{a}})Zhang, Su, Min, Miao, Hu, Fu, Shi, and Chen}]{lzhang_aaai-2023}
Liang Zhang, Jinsong Su, Zijun Min, Zhongjian Miao, Qingguo Hu, Biao Fu, Xiaodong Shi, and Yidong Chen. 2023{\natexlab{a}}.
\newblock Exploring self-distillation based relational reasoning training for document-level relation extraction.
\newblock In \emph{AAAI 2023}.

\bibitem[{Zhang et~al.(2023{\natexlab{b}})Zhang, Zhou, Meng, Su, Chen, and Zhou}]{lzhang_emnlp-2023}
Liang Zhang, Chulun Zhou, Fandong Meng, Jinsong Su, Yidong Chen, and Jie Zhou. 2023{\natexlab{b}}.
\newblock Hypernetwork-based decoupling to improve model generalization for few-shot relation extraction.
\newblock In \emph{EMNLP 2023}.

\bibitem[{Zhang et~al.(2018)Zhang, Qi, and D.~Manning}]{zhang-2018}
Yuhao Zhang, Peng Qi, and Christopher D.~Manning. 2018.
\newblock Graph convolution over pruned dependency trees improves relation extraction.
\newblock In \emph{EMNLP 2018}.

\bibitem[{Zhang et~al.(2019)Zhang, Han, Liu, Jiang, Sun, and Liu}]{Zhangzhen-2019}
Zhengyan Zhang, Xu~Han, Zhiyuan Liu, Xin Jiang, Maosong Sun, and Qun Liu. 2019.
\newblock Ernie: Enhanced language representation with informative entities.
\newblock In \emph{ACL 2019}.

\end{thebibliography}
\bibliographystyle{acl_natbib}

\appendix


\section{Data preprocessing}
\label{A.1}

\begin{figure*}[ht]
\centering
\includegraphics[width=0.8\textwidth]{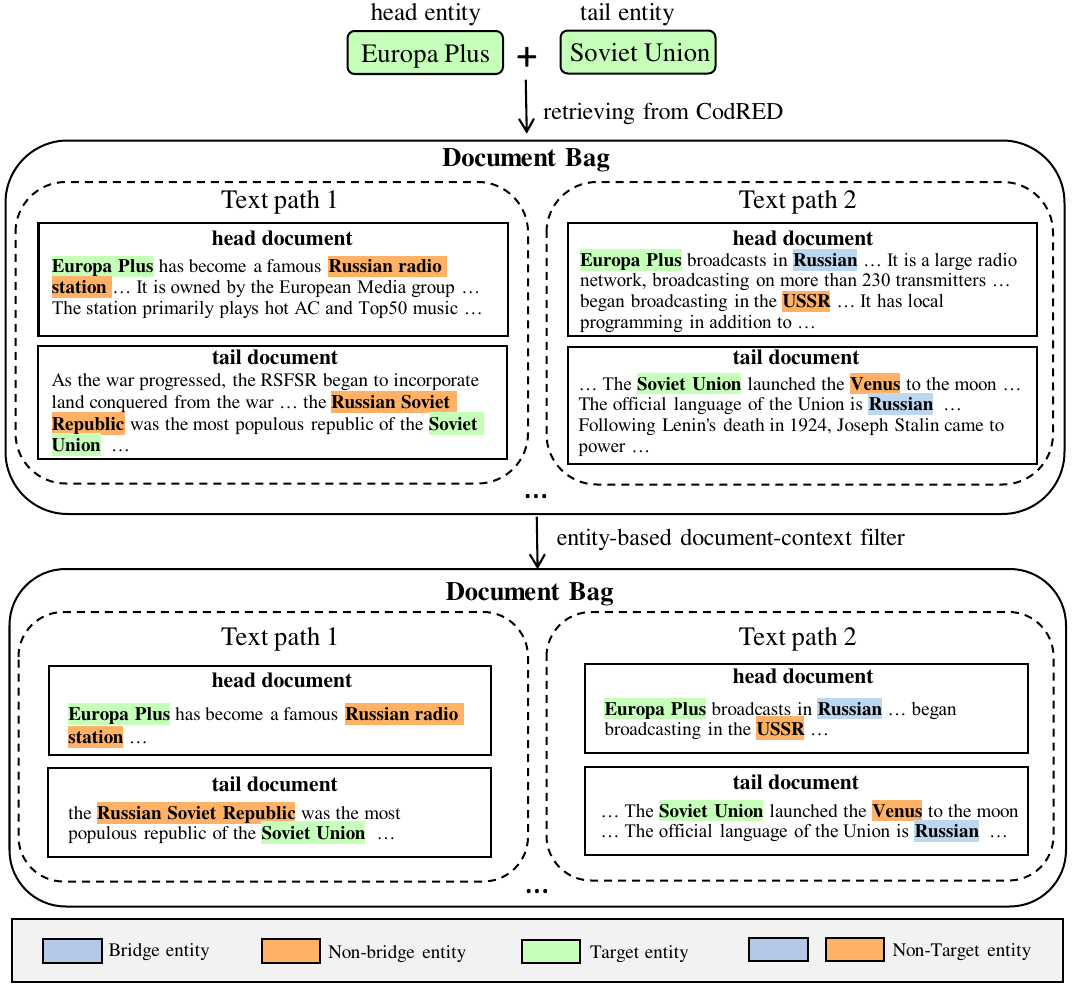}
    \caption{An example of data preprocessing.}
    \label{img:preprocess}
\end{figure*}

Strictly following \citet{wang-2022}, we preprocess our experimental datasets. As shown in Fig \ref{img:preprocess}, we first retrieve relevant text paths for the target entities from the CodRED dataset, forming a document bag. Note that, since the length of each document in a text path may exceed the limit of BERT, we then use an entity-based document-context filter \citep{wang-2022} to select salient sentences for each document and ensure the total length of a text path is less than 512. Finally,  we concatenate the head document and tail document from each text path and then obtain the input of BERT, where the number of text paths in a document bag is the batch size.

\section{Effect of the threshold $\eta$ for semantic-related edge}
\label{A.2}
\begin{figure}[ht]
\centering\includegraphics[width=0.45\textwidth,height=0.25\textwidth]{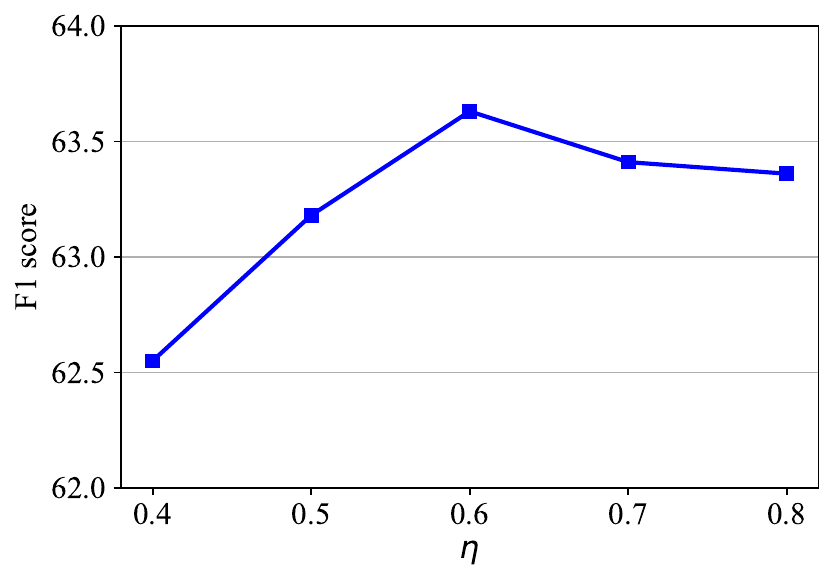}
    \caption{The performance of our model (F1 score) on the development set under the closed setting, using different $\eta$.}
    \label{img:eta}
\end{figure}
We also investigate the impact of the hyper-parameter
$\eta$ on the development set under the closed setting. To this end, we gradually vary $\eta$ from 0.4 to 0.8 with an increment of 0.1 in each step. As shown in Figure \ref{img:eta}, we find that our model achieves the best performance when $\eta$=0.6. Therefore, we set $\eta$=0.6 for all experiments.




\section{Effect of the BERT version}
\label{A.6}
\begin{table}[ht]
\resizebox{.5\textwidth}{!}{
\normalsize
\begin{tabular}{lccccr}
\toprule[1pt]

{\multirow{3}{*}{$\quad$Model}}
& \multicolumn{2}{c}{\textbf{Closed-Test}}         
& \multicolumn{2}{c}{\textbf{Open-Test$\quad$}}    
\\ \cmidrule(l){2-3} \cmidrule(l){4-5} &
\textbf{F1}         & \textbf{AUC}         & \textbf{F1} & \textbf{AUC$\quad$}                  
\\ \midrule
{$\quad$LGCR-BERT$_{base}$} & {61.08} & {60.75}&  {53.45} & {50.15$\quad$} \\
{$\quad$LGCR-BERT$_{large}$} & {62.16} & {61.51} & {54.34} & {51.07$\quad$} \\
\midrule{$\quad$Ours-BERT$_{base}$}
& {64.41} & {66.23}  & {56.68} & {55.87$\quad$}\\
{$\quad$Ours-BERT$_{large}$} & {\textbf{65.27}} & {\textbf{66.98}} & {\textbf{57.44}} & {\textbf{56.43}$\quad$}
\\ \toprule[1pt]
\end{tabular}}
\caption{Experimental results with previous SOTA model, using different BERT versions.}
\label{tab:cross_large}
\end{table}
\begin{table}[ht]
\centering
\resizebox{.4\textwidth}{!}{
\normalsize
\begin{tabular}{lccr}
\toprule[1pt]
    \multirow{2}{*}{Model} & 
    \multicolumn{1}{c}{\textbf{Closed-Dev}} & 
    \multicolumn{1}{c}{\textbf{Open-Dev}} & \\
    \cline{2-4} &
    \textbf{micro-F1} & \textbf{micro-F1}& \\
\hline
{LGCR-BERT$_{base}$} & {82.07} & {81.79}\\
{LGCR-BERT$_{large}$} & {83.12} & {82.65}\\
\hline
{GPT-3.5-turbo} & {28.05} & {25.31}\\
{InstructUIE} & {70.34} & {64.05}\\
{InstructUIE-FT} & {80.72} & {79.66}\\
\hline
{Ours-BERT$_{base}$} & {84.35} & {83.92}\\
{Ours-BERT$_{large}$} & {\textbf{85.18}} & {\textbf{84.60}}\\
\toprule[1pt]
\end{tabular}}
\caption{Experimental results with LLMs, using different BERT versions.}
\label{tab:llm_large}
\end{table}
We also investigate the impact of the BERT version. As shown in Table \ref{tab:cross_large} and Table \ref{tab:llm_large}, when using BERT$_{large}$, our model still significantly outperforms both the previous SOTA: LGCR and LLM-based methods: GPT-3.5-turbo and InstructUIE-FT, demonstrating the effectiveness of our model.

\begin{table}[t]
\small
\renewcommand{\arraystretch}{2}
\centering
\resizebox{0.5\textwidth}{!}{
\begin{tabular}{c p{0.6\linewidth} p{0.6\linewidth}}

\toprule[1pt]
& \multicolumn{1}{c}{GPT-3.5-turbo} & \multicolumn{1}{c}{InstructUIE} \\
\hline
\multirow{3}{*}{Prompt 1} & 
Given the list of relations: [``highway system'', ``country'', ``place of birth'', ...], read the given text path [text path] and predict the relation between head entity: [h] and tail entity: [t]. Answer in the format [``relation label'', ``confidence score(Decimal between 0-1)''] without any explanation. &
Text: [text path]. In the above text, what is the relationship between [head entity] and [tail entity]? Option: [``highway system'', ``country'', ``place of birth'', ...]. Answer: \\
\hline
\multirow{3}{*}{Prompt 2} &
Analyze the information provided in the text path [text path], which includes mentions of the head entity [h] and the tail entity [t]. Infer the relation from the given relation set: [``highway system'', ``country'', ``place of birth'', ...] between the target entity pair. Please consider the entire path and answer in the format [``relation'', ``confidence score(Decimal between 0-1)''] without any explanation. &
Text: [text path]. Find the relationship between [head entity] and [tail entity] in the above text. Option: [``highway system'', ``country'', ``place of birth'', ...]. Answer: \\
\hline
\multirow{3}{*}{Prompt 3} &
Given a text path [text path] containing mentions of the head entity [h] and the tail entity [t] in separate sentences, your goal is to identify and infer the relation from the pre-defined set [``highway system'', ``country'', ``place of birth'', ...] between the target entity pair. Pay close attention to the specific information provided in the path. Answer in the format [``relation'', ``confidence score(Decimal between 0-1)''] without any explanation. &
Text: [text path]. Given the above text, please tell me the relationship between [head entity] and [tail entity]. Option: [``highway system'', ``country'', ``place of birth'', ...]. Answer: \\
\toprule[1pt]

\end{tabular}}

\caption{Zero-shot prompts for GPT-3.5-turbo and InstructUIE on CoRE task}
\label{tab:both}
\end{table}

\section{Experimental settings for LLMs}
\label{A.5}
The detailed experimental settings for the LLMs are as follows:
\begin{itemize}
    \item GPT-3.5-turbo. As implemented in previous studies \cite{han-2023,wang-2023,yuan-2023}, we devise three diverse prompts as shown in Table \ref{tab:both}, which prompt LLM to generate both the relation label and the corresponding confidence scores. Subsequently,  to obtain the bag-level relation prediction, we select the highest confidence score among all text paths within a document bag. 
    We conduct experiment under each prompt separately and then average the results under all prompts as the final experimental result.
    \item InstructUIE and InstructUIE-FT.  We first use LoRA \citep{hu-2021} to obtain InstructUIE-FT. During this process, we set the learning rate to 5e-5 and the batch size to 8. As for the prompt, we follow \citet{wang-2023} to design three prompts as shown in Table \ref{tab:both}. Note that, to obtain the bag-level prediction, we first get the prediction score of the relation label generated by the model. After that, we take the relation label with the highest prediction score among all text paths as the final prediction of a document bag. Like GPT-3.5-turbo, we also average the results under all prompts as the final experimental result.
    
\end{itemize}

\begin{table*}[tp]\small
\centering
\begin{tabular}{p{0.6\linewidth} ccc}

\toprule[1pt]
\hfil \multirow{2}{1.5cm}{Text path}  & \hfil \multirow{2}{*}{Ecrim}  & \hfil \multirow{2}{1.5cm}{Ours w/o $\overline{y}_{rela}$, $\overline{y}_{bias}$} & \hfil \multirow{2}{*}{Ours}
\\\\
\hline

\multicolumn{1}{m{0.6\linewidth}}{\textbf{Text path 1} [head document] \colorbox[cmyk]{0.00, 0.26, 0.53, 0.04}{\textbf{Merovingian script}}, is named after an abbey in \colorbox[cmyk]{0.00, 0.26, 0.53, 0.04}{\textbf{Western France}}, the \colorbox[cmyk]{0.17, 0.00, 0.24, 0.01}{\textbf{Luxeuil Abbey}}, founded by the Irish missionary \colorbox[cmyk]{0.00, 0.26, 0.53, 0.04}{\textbf{St Columba}}  ca.590 ... [tail document]: ...The \colorbox[cmyk]{0.17, 0.00, 0.24, 0.01} {\textbf{Catholic Encyclopedia}} (1913) concludes that the \colorbox[cmyk]{0.00, 0.26, 0.53, 0.04}{\textbf{Salome}} of \colorbox[cmyk]{0.00, 0.26, 0.53, 0.04}{\textbf{Mark}} 15: 40 is probably identical with the mother of the sons of \colorbox[cmyk]{0.00, 0.26, 0.53, 0.04}{\textbf{Zebedee}} in \colorbox[cmyk]{0.00, 0.26, 0.53, 0.04}{\textbf{Matthew}}; the latter is also mentioned in \colorbox[cmyk]{0.00, 0.26, 0.53, 0.04}{\textbf{Matthew}} 20:20 ...}
& \hfil NA \ding{55} & \makecell[c]{described \ding{51}\\by } & \hfil \makecell[c]{described \ding{51}\\by }\\

\hline

\multicolumn{1}{m{0.6\linewidth}}{\textbf{Text path 2}\;\;[head document]: ... \colorbox[cmyk]{0.00, 0.26, 0.53, 0.04}{\textbf{Battambang}} is the capital city of  \colorbox[cmyk]{0.00, 0.26, 0.53, 0.04}{\textbf{Battambang}} province founded in the 11th century, \colorbox[cmyk]{0.00, 0.26, 0.53, 0.04}{\textbf{Lao Thai}}, and \colorbox[cmyk]{0.18, 0.09, 0.00, 0.08}{\textbf{Chinese}} ... The city is situated on the \colorbox[cmyk]{0.17, 0.00, 0.24, 0.01}{\textbf{Sangkae River}} ... [tail document]: ... Textile and garment factories were built by  \colorbox[cmyk]{0.18, 0.09, 0.00, 0.08}{\textbf{Chinese}} investors, and the railway line was extended to \colorbox[cmyk]{0.00, 0.26, 0.53, 0.04}{\textbf{Poipet}} ... Here are some facts and trivia. On the map of the world, \colorbox[cmyk]{0.17, 0.00, 0.24, 0.01}{\textbf{Asia}} terminated in its southeastern point in a cape ...}
& \hfil NA \ding{55} & \hfil NA \ding{55} & \hfil continent \ding{51}\\

\hline

\toprule[1pt]
\end{tabular}
\caption{Two text paths with predicted results sampled from the test set of CodRED dataset.  We use the same style to mark the text paths, where the \colorbox[cmyk]{0.17, 0.00, 0.24, 0.01}{\textbf{target entities}}, \colorbox[cmyk]{0.18, 0.09, 0.00, 0.08}{\textbf{bridge entities}}, and \colorbox[cmyk]{0.00, 0.26, 0.53, 0.04}{\textbf{non-bridge entities}} are marked in green, blue, and orange respectively. In text path 1, with the help of non-bridge entities, both our model and ours w/o $\overline{y}_{rela}$, $\overline{y}_{bias}$ can predict the correct relation. In text path 2, we can find the proposed debiasing strategy helps our model for relation prediction.
}
\label{tab:case}
\end{table*}
\end{document}